\documentclass{article}

\usepackage[preprint]{neurips_2025}

\usepackage[utf8]{inputenc} 
\usepackage[T1]{fontenc}    
\usepackage{hyperref}       
\usepackage{url}            
\usepackage{booktabs}       
\usepackage{amsfonts}       
\usepackage{nicefrac}       
\usepackage{microtype}      
\usepackage{xcolor}         

\usepackage{bm}
\usepackage{graphicx}
\usepackage{amsmath}
\usepackage{amssymb}
\usepackage{mathtools}
\usepackage{amsthm}
\usepackage{physics}

\title{Efficient Deep Learning with\\ Decorrelated Backpropagation}

\author{
  Sander Dalm, Joshua Offergeld, Nasir Ahmad, Marcel van Gerven\\
\textit{Department of Machine Learning and Neural Computing} \\
\textit{Donders Institute for Brain, Cognition and Behaviour} \\
\textit{Radboud University, Nijmegen, the Netherlands}
}

\begin{document}

\maketitle

\begin{abstract}
 The backpropagation algorithm remains the dominant and most successful method for training  deep neural networks (DNNs). At the same time, training DNNs at scale comes at a  significant computational cost and therefore a high carbon footprint. Converging evidence suggests that input decorrelation may speed up deep learning. However, to date, this has not yet translated into substantial improvements in training efficiency in large-scale DNNs. This is mainly caused by the challenge of enforcing fast and stable network-wide decorrelation. Here, we show for the first time that much more efficient training of deep convolutional neural networks is feasible by embracing decorrelated backpropagation as a mechanism for learning. To achieve this goal we made use of a novel algorithm which induces network-wide input decorrelation using minimal computational overhead. By combining this algorithm with careful optimizations, we achieve a more than two-fold speed-up and higher test accuracy compared to backpropagation when training several deep residual networks. This demonstrates that decorrelation provides exciting prospects for efficient deep learning at scale. 

\end{abstract}

\section{Introduction}
Modern AI relies heavily on deep learning (DL), which refers to the training of very deep neural network (DNN) models using massive datasets deployed on high-performance compute clusters~\citep{LeCun2015}. The established way of implementing learning in artificial neural networks is through the backpropagation (BP) algorithm~\citep{Linnainmaa1970,Werbos1974}.
BP is a gradient-based method that implements reverse-mode automatic differentiation to compute the gradients needed for parameter updating in neural networks~\citep{baydin2018automatic}.

At the same time, training of DNNs consisting of many layers of nonlinear transformations is computationally expensive and has been associated with significant energy consumption at a global scale~\citep{Garcia-Martin2019,Vries2023,thompson2021deep,Luccioni2022,strubell2020energy,Luccioni2023}. 
Hence, we are in need of green AI solutions which can significantly reduce the energy consumption of modern AI systems~\citep{Gibney2022,Schwartz2020, Wange}.
Previous work describes several ways to reduce the carbon consumption of deep learning, focusing mostly on more effective deployment methods~\citep{Patterson2022}.  
Another key reason for deep learning's large carbon consumption, however, is the inefficiency of the backpropagation algorithm itself. BP requires many iterations of gradient descent steps until parameters converge to their optimal values. For example, training a large GPT model can easily take several weeks on a large compute cluster~\citep{Brown2020}.

In this paper, we show that the efficiency of deep learning can be substantially increased by enforcing that layer inputs remain decorrelated throughout the network. 
Decorrelation has previously been proposed to make credit assignment more efficient in neural networks~\citep{lecun2002efficient}.
Intuitively, if a network layer's inputs have highly correlated features, it will be more difficult for the learning algorithm to perform credit assignment as it is now unclear if a change in the loss should be attributed to feature $i$ or feature $j$ in case both are correlated.
Previous work has indeed shown that promoting decorrelation positively impacts training in relatively shallow networks~\citep{Ahmad2022}. This advantage of decorrelation can be theoretically understood as a way to better align the gradient update with that of the natural gradient~\citep{Ahmad2024,Amari1998}.
Similar results have been obtained when inputs are whitened such that inputs are not only decorrelated but also forced to have unit variance~\citep{luo}. Whitening has also been shown to have a positive impact in training deep networks, resulting in faster convergence when measured by number of epochs~\citep{Huangi2018}.

However, since decorrelation and whitening methods themselves incur computational overhead, convergence speedup as a function of the number of epochs is not reflected in reduction in wall-clock time or carbon consumption, which is the metric of choice in green AI. Furthermore, other work indicates that whitening may have a  convergence speed that is similar to that of batch-normalized networks~\citep{Desjardins2015} and generalization performance may be negatively impacted~\citep{wadia2021whitening}. 
Hence, while the theoretical benefits of decorrelation and whitening are encouraging, their practical application to significantly reduce real-world training time in modern deep neural networks remains to be shown.

In this paper we demonstrate, for the first time, that training of DNNs can be substantially improved in practice by employing a novel decorrelated backpropagation (DBP) algorithm.
DBP combines automatic differentiation with an efficient iterative local learning rule which effectively decorrelates layer inputs across the network. This iterative decorrelation procedure generalizes a method which was previously introduced in the context of biologically plausible learning, demonstrating faster convergence in limited-depth fully-connected neural networks~\citep{Ahmad2022}.
Here, we extend the method to allow efficiently and effectively training of deep convolutional neural networks~\citep{He2016deep}. 

Our method is compared on four DNN architectures trained on the ImageNet dataset. To ensure a fair comparison across methods, we include batch normalization and optimized hyper-parameter optimization for both methods separately.

We show that wall-clock time of DNN training can be significantly reduced, up to 50\% of that of regular BP, also reducing carbon emissions. Furthermore, we show that test accuracy of DBP-trained networks is consistently higher than that of BP-trained networks. Overall, our results show that including a local decorrelation mechanism across network layers can substantially improve training of deep neural networks.

\section{Methods}

\subsection{Decorrelated backpropagation}

Let us consider a DNN consisting of $K$ layers implementing some parameterized nonlinear transform. Deep learning typically proceeds via reverse-mode automatic differentiation, also known as the backpropagation (BP)  algorithm. Backpropagation updates the parameters $\bm{\theta}$ (weights and biases) of a network according to
\begin{equation}
\bm{\theta} \gets \bm{\theta} - \eta \nabla_{\bm{\theta}} \mathcal{L}
\end{equation}
where $\eta$ is the learning rate, $\bm{\theta}$ are the network parameters and $\nabla_{\bm{\theta}} \mathcal{L}$ is the gradient of the loss. 

Decorrelated backpropagation (DBP) operates as regular backpropagation but additionally enforces the input to network layers to be decorrelated. 
Since decorrelation is imposed independently in each layer, we concentrate on describing the decorrelation procedure for one such layer, implementing a nonlinear transformation
\begin{equation}
\vb{y} = f\left(\vb{W}\vb{x}\right)
\label{eq:layer}
\end{equation}
with input $\vb{x} \in \mathbb{R}^d$, weights $\vb{W} \in \mathbb{R}^{m \times d}$ and output $\vb{y} \in \mathbb{R}^m$. The nonlinear transformation may be either the transformation in a fully-connected layer or the kernel function of a convolutional layer applied to a (flattened) input patch. We ignore biases without loss of generality since these can be represented using weights with fixed constant input.

Input decorrelation refers to the property that the second-moment matrix $\langle \vb{x} \vb{x}^\top \rangle$ is diagonal for inputs $\vb{x}$, where the expectation is taken over the data distribution. In case of whitened inputs,  $\langle \vb{x} \vb{x}^\top \rangle$ is additionally assumed to be the identity matrix~\citep{Kessy2015}. We will refer to the former as the covariance constraint and the latter as the variance constraint.
To ensure that the input $\vb{x}$ is decorrelated, we assume that it is the result of a linear transform
$
\vb{x} = \vb{R} \vb{z}
$, where the decorrelating matrix $\vb{R}$ transforms the correlated input $\vb{z}$ into a decorrelated input $\vb{x}$. Hence, the transformation in a layer is described by
\begin{equation}
    \vb{y} = f(\vb{W}\vb{R}\vb{z}) = f(\vb{W}\vb{x}) = f(\vb{A}\vb{z})
\end{equation}
with $\vb{A} = \vb{W}\vb{R}$. The output $\vb{y}_l$ of the $l$-th layer becomes the correlated input $\vb{z}_{l+1}$ to the next layer. 

\subsection{Decorrelation learning rule}

To implement decorrelated backpropagation, we need to update individual decorrelation matrices for each of the layers in the network.  This can be achieved by minimizing $K$ decorrelation loss functions {\em in parallel} with minimizing the BP loss with the aim of accelerating learning speed.
One way to achieve whitening as a restricted form of decorrelation is via established approaches such as zero-phase component analysis (ZCA)~\citep{Bell1995,Bell1997-qz}. 
However, as described in Appendix~\ref{sec:zca}, naive application of ZCA is prohibitively costly in practice since it requires computing an expensive ZCA transform for the inputs of each layer at each gradient-descent step.

To enable efficient decorrelation and whitening in DNNs, we generalize a recently introduced approach that allows for efficient network-wide decorrelation~\citep{Ahmad2022}.

To derive the decorrelation learning rule, we start by defining the total decorrelation loss as the average decorrelation loss across layers. The layer-specific decorrelation loss is defined as $L = \langle \ell(\vb{x}) \rangle$, where the expectation is taken over the empirical distribution and 
$
\ell(\vb{x}) = \sum_{i=1}^d \ell_i(\vb{x})
$
is the sum over unit-wise losses $\ell_i$. We define the unit-wise loss as
\begin{equation}
\ell_i(\vb{x}) = (1-\kappa) \frac{1}{2} \sum_{j\neq i} (x_ix_j)^2 + \kappa \frac{1}{4} (x_i^2 - 1)^2
\label{eq:singleloss}
\end{equation}
where the first term is the zero covariance constraint imposed by decorrelation and the second term is the unit variance constraint imposed by whitening. The whitening parameter $\kappa \in [0,1]$ trades off between the covariance constraint ($\kappa=0$) and the variance constraint ($\kappa=1$) and is whitening inducing when $\kappa > 0$. 
we may now write the layer-wise decorrelation loss compactly as 
\begin{equation}
\ell(\vb{x}) = \left\langle (1-\kappa) \textrm{Tr}(\vb{C} \vb{C}^\top)  + \kappa  \textrm{Tr}(\vb{V} \vb{V}^\top) \right\rangle
\end{equation}
where $\vb{C} = \vb{x} \vb{x}^\top - \vb{D}$ with $\vb{D}=\textrm{diag}(x_1^2,\ldots,x_d^2)$ and $\vb{V} = \textrm{diag}(x_1^2-1,\ldots,x_d^2-1)$.

Since the decorrelation loss is minimized independently in each layer, we concentrate on describing the minimization of the decorrelation loss for one such layer, as in Eq.~\ref{eq:layer}.
Let us first consider decreasing the correlation between variables for one input vector $\vb{x}$. This can be achieved using an update step of the form 
\begin{equation}
\vb{x} \gets \vb{x} - \epsilon \nabla_{\vb{x}} \ell
\label{eq:ustep}
\end{equation}
with $\epsilon$ the decorrelation learning rate. To compute the gradient, we first consider the partial derivative of the unit-wise loss with respect to the input $x_i$ given by
\begin{equation}
     \frac{\partial \ell_i}{\partial x_i} = (1-\kappa) \sum_{j \colon j \neq i} \left( x_i x_j \right) x_j + \kappa (x_i^2 - 1)x_i 
\end{equation}
which, when vectorized across inputs, yields 
$
\nabla_{\vb{x}} \ell = (1-\kappa) \vb{C}\vb{x} + \kappa \vb{V}\vb{x}
$.
If we now plug this into the update step in Eq.~\ref{eq:ustep}, we obtain
$
\vb{x} \gets \vb{x} - \epsilon \left( (1-\kappa) \vb{C} + \kappa \vb{V}\right) \vb{x}
$. Note, however, that our goal is to derive an update rule for $\vb{R}$ rather than $\vb{x}$. This can be obtained using the identity $\vb{x} = \vb{R}\vb{z}$ since this allows us to write
$
\vb{R}\vb{z} \gets \vb{R}\vb{z} - \epsilon \left( (1-\kappa) \vb{C} + \kappa \vb{V}\right) \vb{R}\vb{z}
$. 
Dividing by $\vb{z}$ on both sides and averaging over input samples we obtain the decorrelation learning rule
\begin{equation}
\label{eq:decorrule}
\vb{R} \gets \vb{R} -  \epsilon \left\langle (1-\kappa) \vb{C} + \kappa \vb{V}\right\rangle \vb{R} \,.
\end{equation}
This decorrelation rule allows for efficient batch updating of $\vb{R}$ since it only requires computing the decorrelated input covariance followed by multiplication with the decorrelation matrix. 

\begin{figure*}[!ht]
\centering
\includegraphics[width=\textwidth]{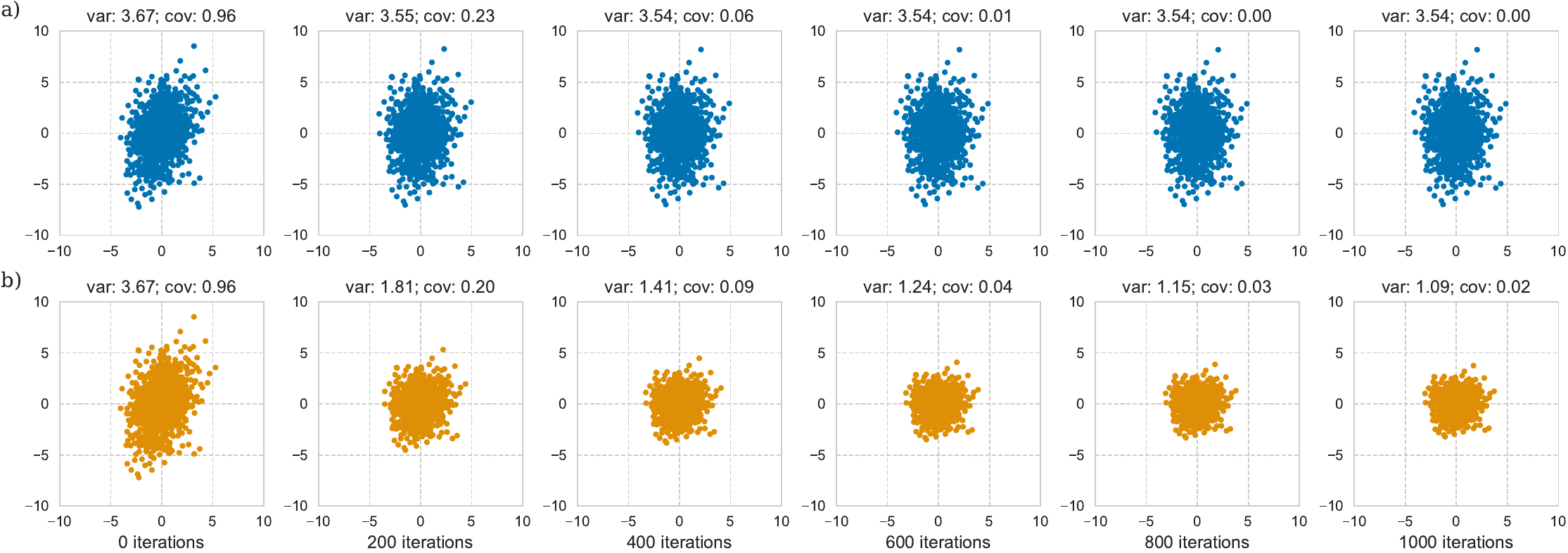}
\caption{Demonstration of the decorrelation rule on correlated input data consisting of 1000 examples and two covariates with decorrelation learning rate $\epsilon = 0.001$. a) Decorrelation using $\kappa=0$. b) Whitening using $\kappa=0.5$. Mean variance and covariance reported for different iterations. }
\label{fig:iterations}
\end{figure*}

Figure~\ref{fig:iterations} illustrates effective decorrelation and whitening on two-dimensional input data using Eq.~\eqref{eq:decorrule}. 

\subsection{Decorrelating deep convolutional neural networks}

Effective training of DNNs using DBP requires several modifications that improve algorithm stability, learning speed and allow application to convolutional rather than fully connected layers. 
 As a canonical example, we consider a deep residual network (ResNet). A ResNet consists of multiple residual blocks  that implement modern network components such as convolutional layers and residual connections~\citep{He2016deep}. 
In the following, we describe the modifications that are required for effectively applying DBP in deep neural networks.

\begin{figure*}[!ht]
\centering
\includegraphics[width=\textwidth]{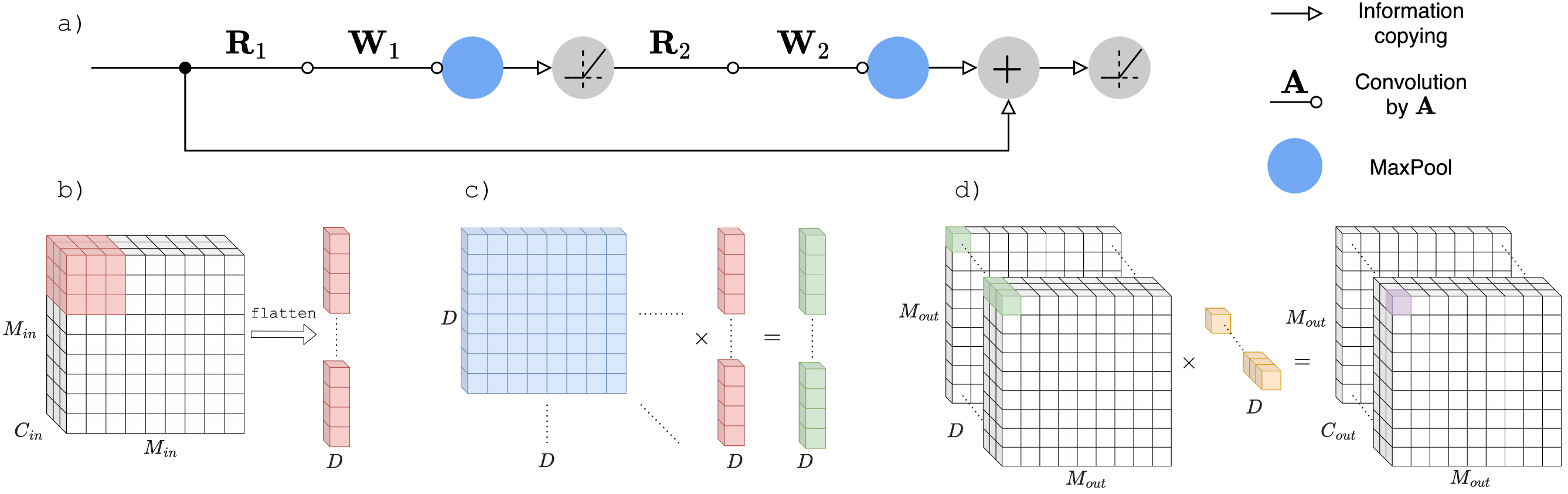}
\caption{Implementation of decorrelated backpropagation in residual networks. a) Residual blocks as implemented by our networks. b) Patch-wise flattening of the inputs with a flattened dimension of $d=M \times M \times C_\text{in}$. c) Decorrelating/whitening transform of the data by decorrelation matrix $\vb{R}$. d) 1 $\times$ 1 convolution operation with the weights $\vb{W}$ on the decorrelated input patches.}
\label{fig:resnet}
\end{figure*}

\paragraph{Decorrelating convolutional layers}
We apply the decorrelation learning rule to convolutional layers as follows. Consider a convolutional layer with input size $S = M_\text{in} \times M_\text{in} \times C_\text{in}$, where $M_\text{in} \times M_\text{in}$ is the size of the feature map and $C_\text{in}$ is the channel dimension. Naive implementation of decorrelation would require updating a $S \times S$ decorrelation matrix $\vb{R}$, which is computationally prohibitive in case of large feature maps. Instead, as shown in Fig.~\ref{fig:resnet}, we do not decorrelate the layer's entire input, but only the local image patches, so that for an image patch with dimensionality $D = K \times K \times C_\text{in}$, matrix $\vb{R}$ would be of much smaller size $D \times D$. The output of this patchwise decorrelation operation would then be $D$ for every image patch, after which we can apply a $1 \times 1$ convolutional kernel of size $D \times C_\text{out}$ for the forward pass. This local decorrelation is sufficient to ensure more efficient learning of the kernel weights. Figure~\ref{fig:resnet} depicts the structure of a residual block that is extended to implement decorrelating transforms.

\paragraph{Subsampling during covariance estimation}
Second, applying the decorrelation learning rule can become costly when averaging over input samples in Eq.~\ref{eq:decorrule} in case of convolutional layers. This is due to the need to compute an outer product $\vb{X} \vb{X}^\top$ with $\vb{X} \in \mathbb{R}^{D \times p}$ the input to a layer, where $p$ is the product of the number of batch elements times the number of patches. 
We found empirically that updates of $\vb{R}$ need not use the entire mini-batch to learn the correlational structure of the data. Sampling just 10\% of the samples in each batch yields almost identical performance while significantly reducing the computational overhead of computing $\vb{R}$'s updates, as shown in Appendix~\ref{sec:sampling}.

\paragraph{Efficient matrix products}
A final optimization that decreases the computational overhead of the decorrelating transform is to combine the matrices $\vb{W}$ and $\vb{R}$ into a condensed matrix $\vb{A} = \vb{W}\vb{R}$ prior to multiplying by the correlated input $\vb{z}$. The dimensionality of $\vb{z}$ is much larger than that of $\vb{W}$ due to the batch dimension, and thus we replace an expensive multiplication $\vb{W}(\vb{R}\vb{z})$ by a cheaper multiplication $(\vb{W}\vb{R})\vb{z}$, significantly reducing the time required for performing a forward pass. As an additional benefit, when training is complete, we only need to store the $\vb{A}$ matrices.

\subsection{Experimental validation}
\label{sec:exp}

To evaluate learning performance in DNNs, 18-, 34- and 50-layer ResNet models as well as an AlexNet model \citep{alexnet} were trained using both BP and DBP. Performance was evaluated using the ImageNet Large Scale Visual Recognition Challenge (ILSVRC) dataset~\citep{ImageNet}. This dataset spans 1000 object classes and contains 1,281,167 training images, 50,000 validation images and 100,000 test images.

Data preprocessing consisted of the following steps. First, images were normalized by subtracting the means from the RGB-channels' values and dividing by their standard deviation. Next, images were rescaled to $256 \times 256$ and a $224 \times 224$ crop was taken from the center.  We used no data augmentation, but shuffled the order of the data every epoch.

All models and algorithms were implemented in PyTorch~\citep{NEURIPS2019_9015} and run on a compute cluster (Surf Snellius) using Nvidia A100 GPUs and Intel Xeon Platinum 8360Y processors. 
To initialize the weights of our models, we set $\vb{R}$ to the identity matrix and used He initialization~\citep{he2015delving} for $\vb{W}$.
Models were trained to minimize the categorical cross-entropy loss. Exploratory analysis revealed negligible performance difference as a function of batch size and a batch size of 256 was chosen. Instead of using the full batch of images in each decorrelation update step, we reduce the number of samples to 10\%. We empirically found this value to show negligible loss in performance, while significantly speeding up runtime. See Appendix~\ref{sec:sampling} for an analysis. 
The accuracy reported is the top-1 performance of the models. Reported wall-clock time was measured as the runtime of the training process, excluding performance testing. Total compute time used is estimated at 400 hours, which includes initial exploration, hyper-parameter search and running the large-scale experiments.

For the $\kappa$ parameter, a value of 0.5 was chosen, balancing the covariance and variance constraints. 
For updating $\vb{W}$, we use the Adam optimizer~\citep{Kingma2014} with a learning rate of $\eta = 1.6 \cdot 10^{-4}$ and optimizer parameters $\beta_1 = 0.9$ and $\beta_2 = 0.999$.  We added a small value of $10^{-8}$ to the denominator of the updates for numerical stability.
The decorrelation matrix $\vb{R}$ is updated with stochastic gradient descent using a learning rate of $\epsilon = 10^{-5}$. 
To ensure a fair comparison between algorithms, we made sure that all algorithms attained their highest convergence speed by performing a two-dimensional grid search over the BP ($\eta$) and decorrelation ($\epsilon$) learning rates. See Appendix~\ref{sec:grid} for the grid search results on the ResNet18 model.

\section{Results}

In the following, we analyze the performance of DBP on different DNN models trained on the ImageNet task.

\subsection{DBP effectively decorrelates inputs to all network layers}
\label{sec:decor_losses}

\begin{figure*}[!ht]
\centering
\includegraphics[width=\textwidth]{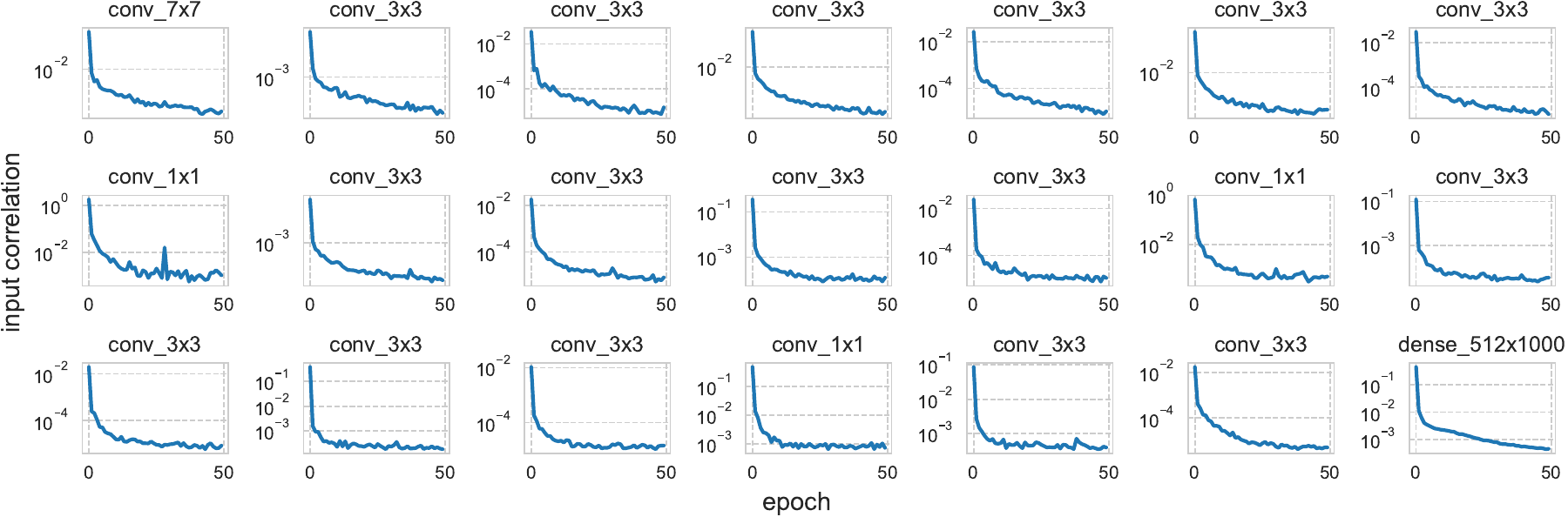}
\caption{Input decorrelation when training a ResNet18 model on ImageNet for 50 epochs. Network layers are ordered from left to right and from top to bottom. Panel titles indicate the layer type.}
\label{fig:decor_loss}
\end{figure*}

We first verified that DBP indeed effectively reduces input decorrelation across layers in DNNs. Figure~\ref{fig:decor_loss} shows the change is input correlation for each layer of the ResNet18 model during the training process for DBP. To quantify input correlation we use the mean squared values of the strictly lower triangular part of the $\vb{C}$ matrix. Layer inputs become rapidly decorrelated during the first 10 epochs and correlations remain consistently low. This demonstrates that the decorrelating learning rule indeed effectively learns to decorrelate the inputs to all network layers.

\subsection{DBP converges much faster than BP}
\label{sec:results_epoch}

\begin{figure*}[!ht]
\begin{center}
\includegraphics[width=0.9\linewidth]{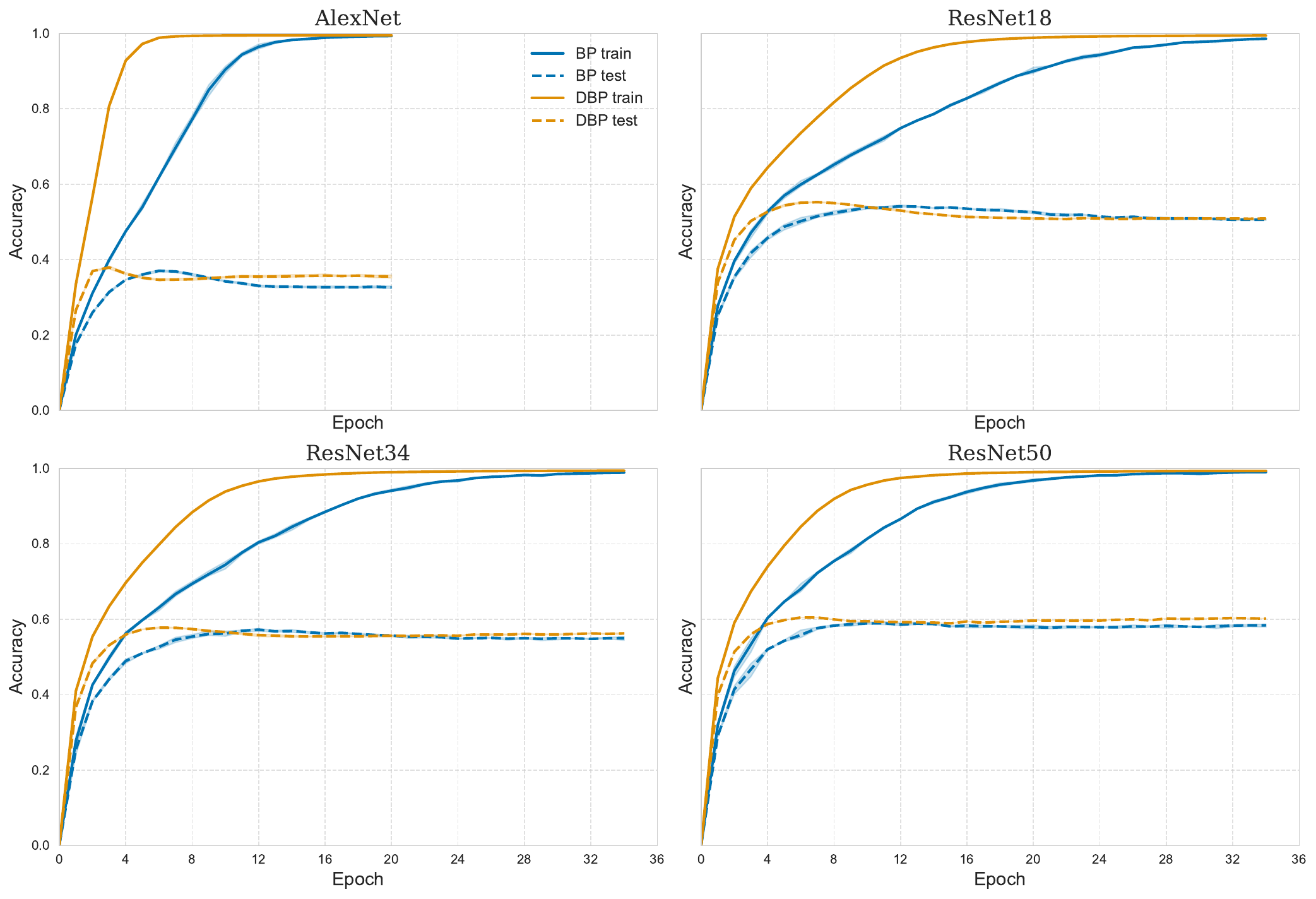}
 \caption{Train and test performance of BP and DBP on Imagenet as a function of epochs for different deep neural network architectures. AlexNet was trained for 20 epochs and the deeper ResNet models were trained for 35 epochs. a) AlexNet. b) ResNet18. c) ResNet34. d) ResNet50.}
\label{fig:architectures_epoch}
\end{center}
\end{figure*}

Figure~\ref{fig:architectures_epoch} compares training and test convergence using DBP and BP as a function of the number of epochs across models. For each model, convergence of both train and test accuracy is much faster for DBP compared to BP, demonstrating the positive impact of decorrelation. Furthermore, test performance is consistently higher for DBP compared to BP. 

\subsection{DBP training yields shorter wall-clock times}
\label{sec:results_wallclock}

\begin{figure*}[!ht]
\begin{center}
\includegraphics[width=0.9\linewidth]{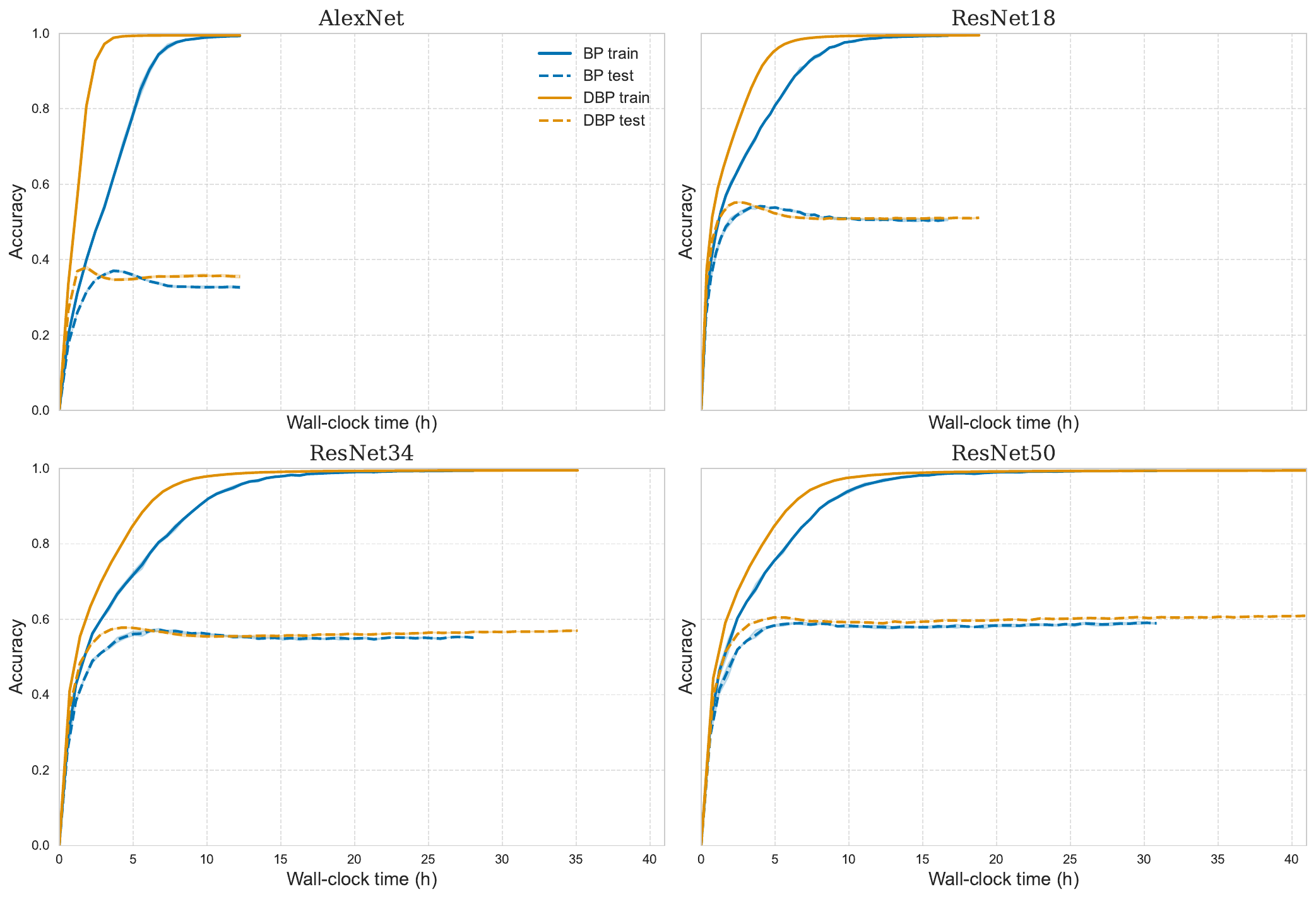}
 \caption{Train and test performance of BP and DBP on Imagenet as a function of wall-clock time for different deep neural network architectures. AlexNet was trained for 20 epochs and the deeper ResNet models were trained for 35 epochs. a) AlexNet. b) ResNet18. c) ResNet34. d) ResNet50.}
\label{fig:architectures_walltime}
\end{center}
\end{figure*}

As emphasized in the Introduction section, speedup in terms of epochs may not necessarily translate into speedup in terms of wall-clock time due to additional computational overhead. Figure~\ref{fig:architectures_walltime} shows that, while the speedup is indeed attenuated due to this overhead, DBP remains faster than BP, reducing training time up to 50\% across models when using peak test performance as a stopping criterion. 

\subsection{DBP reduces carbon emission}

\begin{figure*}[!ht]
\begin{center}
\includegraphics[width=1\linewidth]{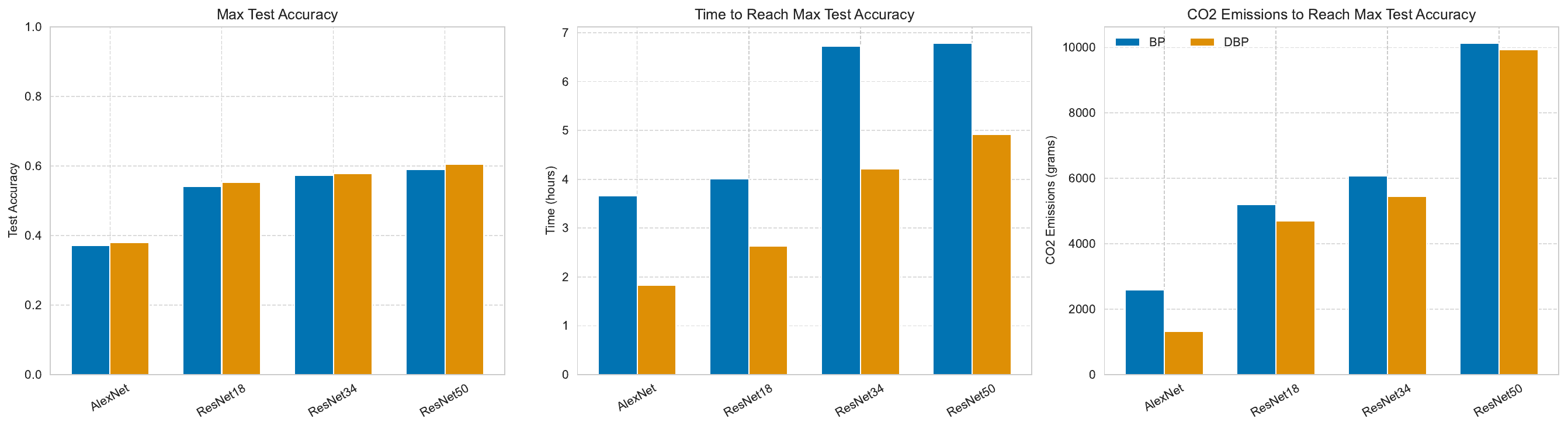}
 \caption{a) Maximum test accuracy b) Time needed to achieve maximum test accuracy c) Estimated reduction in carbon consumption when using DBP over BP. Estimates produced using \url{https://github.com/mlco2/codecarbon}.}
\label{fig:main_results}
\end{center}
\end{figure*}

Reduced wall-clock time does not only reduce the need for compute capacity but also decreases the carbon emission of deep learning. Figure~\ref{fig:main_results} summarizes our results, showing faster learning at higher accuracy with reduced carbon consumption.

\subsection{Impact of whitening}

\begin{figure}[!ht]
\begin{center}
\includegraphics[width=1\textwidth]{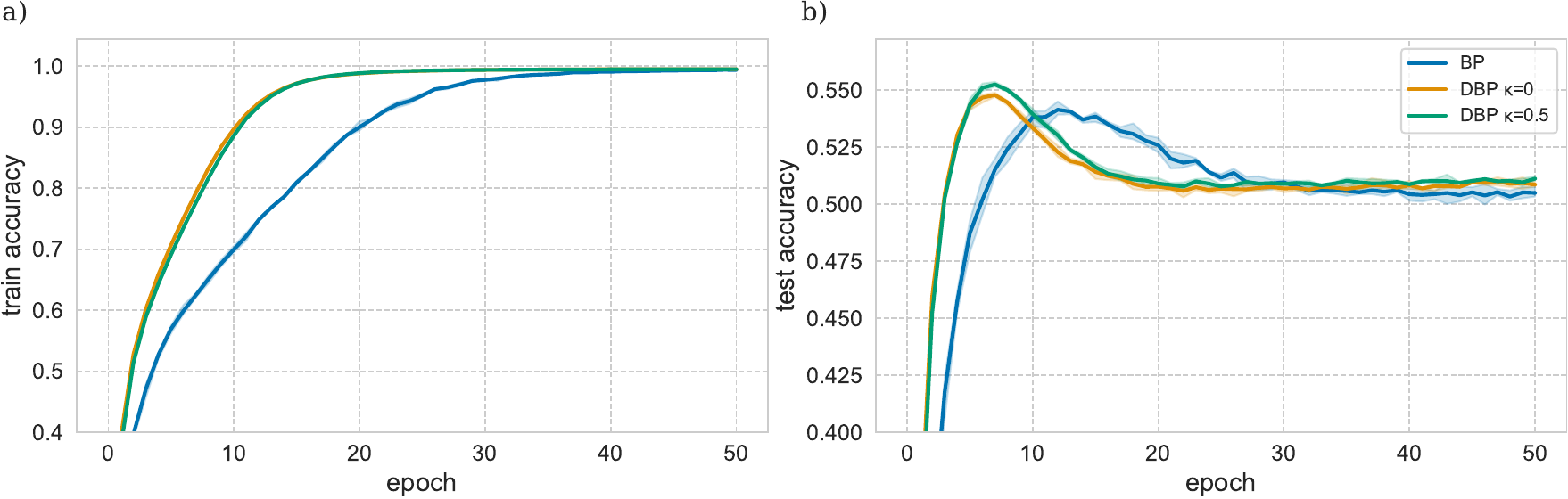}
 \caption{Performance of BP and DBP on ResNet18 using $\kappa=0$ (decorrelation) and $\kappa=0.5$ (whitening). Reported results are the average of three randomly initialized networks, where minimal and maximal value are indicated by the error bars. a) Train accuracy as a function of the number of epochs. b) Test accuracy as a function of the number of epochs.}
\label{fig:result1}
\end{center}
\end{figure}

The question remains how the choice of the whitening parameter $\kappa$ influences performance. To this end, we compared the performance of the ResNet18 model using $\kappa=0$ (decorrelation) and $\kappa=0.5$ (whitening) for three random initializations. Figure~\ref{fig:result1} show that performance is slightly improved for $\kappa=0.5$ compared to $\kappa=0$. However, as shown in Appendix~\ref{sec:kappaeffect}, the optimal choice of $\kappa$ depends on the chosen architecture and dataset. 

\subsection{Robustness to data regimes}

The results in Sections~\ref{sec:results_epoch} and~\ref{sec:results_wallclock} intentionally show performance in a setup without data augmentation or other methods to aid performance, as the aim was to make a clean comparison of the convergence properties of the algorithms. In this section, we will test the robustness of our findings by exposing both BP and DBP to two additional data regimes: data augmentation consisting of random cropping and horizontal flipping as described in \citet{He2016deep}, as well as downsampling the training set to 10\%, stratified over classes.

\begin{figure}[!ht]
\begin{center}
\includegraphics[width=.75\linewidth]{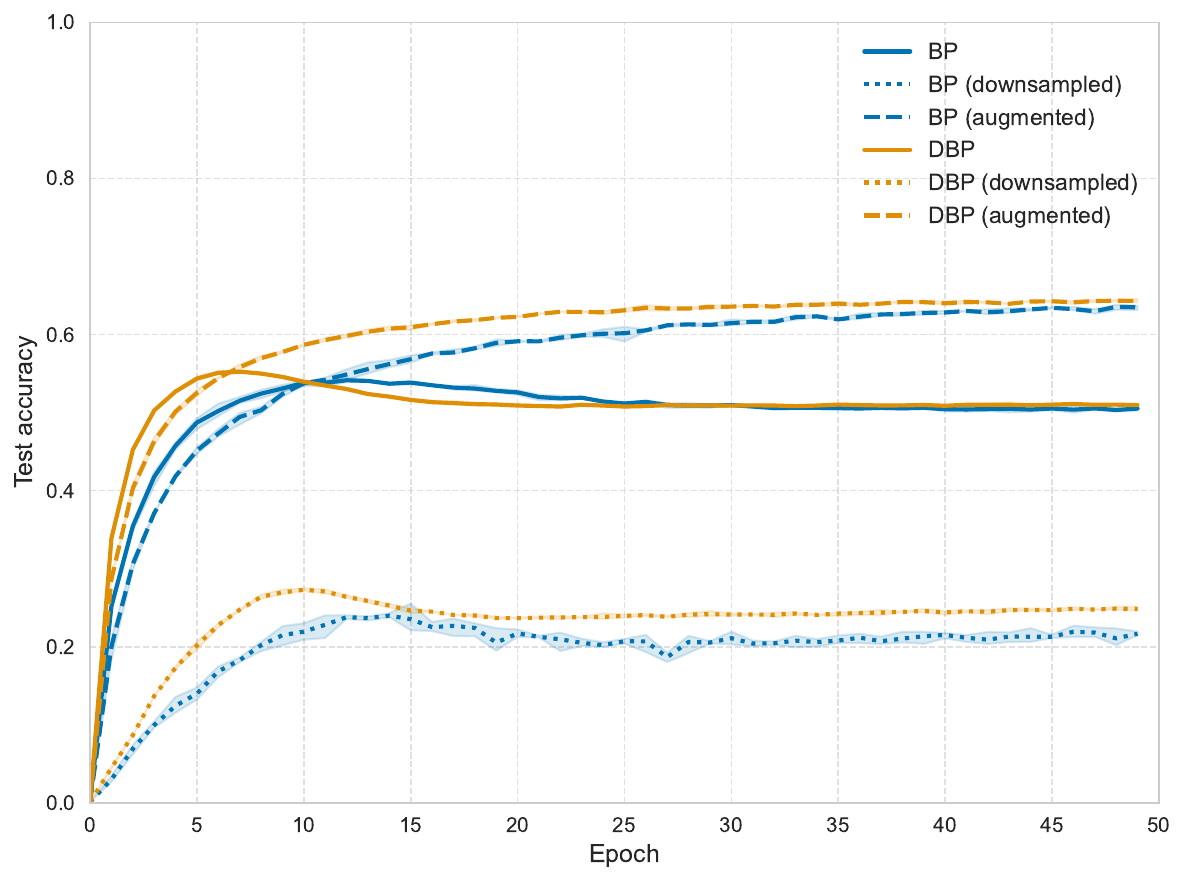}
 \caption{Test performance of BP and DBP ($\kappa=0.5$) on ResNet18 using for the regular ImageNet dataset (solid line), data augmentation (dashed line) and data downsampled to 10\% (dotted line).}
\label{fig:aug_downsample}
\end{center}
\end{figure}

Results in Figure~\ref{fig:aug_downsample} indicate that DBP's outperformance of BP is robust in all three data regimes. Data augmentation improved test performance over baseline for both algorithms, which now reach 65\% top-1 test accuracy, close to published benchmarks \cite{He2016deep}. Data downsampling obviously hurt test performance for both algorithms, but DBP's higher maximum test accuracy and faster convergence remain intact.

\section{Discussion}

Increasing the efficiency of deep learning is of key importance if we are to reduce the carbon footprint of AI. In this paper, using four deep neural network architectures trained on the Imagenet dataset, we have shown that decorrelated backpropagation, which decorrelates the inputs to each network layer, provides a viable path towards more efficient deep learning.
Results show that, by replacing BP with DBP, we achieve higher test accuracy using up to half the number of training epochs, resulting in reduced carbon emission. Results appear to be robust under both data augmentation and data downsampling.

The observed reduction in training time and associated carbon emission of DBP is the result of a complex interplay between having a closer alignment of the gradient with the natural gradient, yielding faster convergence per epoch, and the need to decorrelate inputs across network layers, which comes with additional overhead per epoch. This overhead stems from a number of factors which apply to all $1 \leq k \leq K$ network layers. First, the need to store the decorrelation matrices $\vb{R}_k$. Second, computing the gradient update \eqref{eq:decorrule} for each matrix $\vb{R}_k$ in parallel. Third, additional matrix multiplications $\vb{R}_k$ and $\vb{R}^\top_k$ during the forward and backward pass of the backpropagation algorithm. These factors contribute to increased memory and compute requirements though this is more than compensated for by the much faster convergence of DBP compared to BP.

These considerations suggest that even larger performance gains may be achievable if we can further reduce the memory and compute requirements of DBP. One way to achieve this may be to exploit sparseness structure in the decorrelation matrix. For instance, rather than learning a full matrix $\vb{R}$ we may choose to learn a lower-triangular matrix, which can be achieved by using the strictly lower triangular part of $\vb{C}$ in Eq.~\ref{eq:decorrule}. Another approach may be to use a low-rank approximation of $\vb{R}$. Finally, we may choose to only incorporate decorrelation in that subset of layers which have the biggest impact in terms of reducing convergence time~\citep{Huangi2018}. 
Further speedups may be achievable by improving the decorrelation algorithm itself, for instance, by using layer-specific decorrelation learning rates or subsampling rates that take layer properties into account. 
We expect that additional theoretical work on the optimal alignment of layer inputs combined with efficient low-level implementations will yield even greater gains in convergence speed.

It is important to realize that DBP needs proper fine-tuning. The stability of the algorithm depends on the interplay between the backpropagation learning rate $\eta$ and the decorrelation learning rate $\epsilon$. This ratio needs to be set appropriately such that the decorrelation rate throughout all network layers remains optimal. That is, a too slow decorrelation learning rate will not improve convergence speed while still incurring computational overhead. On the other hand, a too fast decorrelation learning rate will reduce the ability of backpropagation to keep up with changes in input statistics across the network, leading to instabilities. We empirically observe that this fine-tuning depends on the employed dataset, network architecture and loss function. Additional work on the interaction between these factors may provide insights into how to optimally set these parameters. 
Our results further show that performance is maximized for some optimal value of $\kappa$ which most effectively balances the covariance and variance constraints. As a rule of thumb, the setting $\kappa=0.5$ is reasonable though we may also choose to set $\kappa=0$, further simplifying the decorrelation learning rule to $\vb{R} \gets \vb{R} -  \epsilon \left\langle \vb{C} \right\rangle \vb{R}$.

Another important consideration is normalizing the decorrelation update magnitude according to layer size. Note that Eq.~\ref{eq:decorrule} ignores the layer size in the normalization, which implies that decorrelation updates have a larger impact in larger layers whereas whitening induced by $\kappa > 0$ has a smaller impact in larger layers. In simulation work, we observe that such normalization ensures that the impact of decorrelation updates is comparable for different layer sizes. However, in our empirical work using very deep neural networks, we find that disregarding this normalization actually improves performance. 
There are different reasons why this may be the case. First, the resulting stronger decorrelation in deep large layers when ignoring normalization may have a beneficial impact since deeper layers are more strongly affected by multiple preceding nonlinearities. Second, the decorrelation loss that is minimized by Eq.~\ref{eq:decorrule} may not be the optimal metric to ensure fastest convergence. That is, there may be an as yet unidentified discrepancy between the decorrelation gradient direction and the optimal gradient direction to align the inputs.

It should be mentioned that network-wide decorrelation comes with a number of additional advantages, which we did not further explore here. As shown in~\citep{Ahmad2022}, decorrelated layer inputs allow for extremely efficient computation of a optimal linear mapping between arbitrary network layers. That is, when inputs are decorrelated, the matrix that needs to be inverted to compute that ordinary least squares solution becomes diagonal, reducing the inversion from $\mathcal{O}(d^3)$ to $\mathcal{O}(d)$ with $d$ the number of inputs. This property allows for filter visualization in deep networks with applications in explainable AI~\citep{Rudin2019,Ras2022} as well as for network compression at virtually no computational overhead, which may further decrease carbon consumption at inference time~\citep{Wange}. 

Intriguingly, input decorrelation and whitening have also been shown to be a feature of neural processing~\citep{Franke2017-vo,Bell1997-qz,Pitkow2012-wz, Segal2015-gq,Dodds2019-nq,Graham2006-qk,King2013-gz}. The work presented here suggests that decorrelation may be an important factor when considering synaptic plasticity mechanisms, warranting further investigation.

In this work, we demonstrated a significant speedup in deep feedforward neural networks, which are state-of-the-art models for computer vision. In follow-up work, we aim to explore the efficiency gains that can be obtained for other computational tasks and network architectures. Exciting prospects are the use of decorrelated backpropagation for deep reinforcement learning~\citep{Mnih2015,Schrittwieser2020} or training of generative models~\citep{Furfaritony2002,Gu2024,Bommasani2021}, both of which are notoriously resource intensive. Concluding, by demonstrating the efficiency of decorrelated backpropagation in modern deep neural networks, we hope to contribute to reducing the energy consumption of modern AI systems.

\section*{Acknowledgements}
This publication was partly financed by the Dutch Research Council (NWO) via the Dutch Brain Interface Initiative (DBI2) with project number 024.005.022 and a TTW grant with project number 2022/TTW/01389357. We thank Mustafa Acikyurek for his support, SURF (\url{www.surf.nl}) for providing access to the National Supercomputer Snellius and Hewlett Packard for providing a Z8 workstation to support this work.

\bibliography{library.bib}
\bibliographystyle{plainnat} 
\newpage

\appendix

\section{Comparison with zero-phase component analysis}
\label{sec:zca}

Zero-phase component analysis (ZCA)~\citep{Bell1995,Bell1997-qz} is defined by the whitening transform
$
\vb{x} = \vb{C}\vb{x}_c
$, 
where $\vb{x}_c = \bar{\vb{x}} - \bm{\mu}$ with $\bm{\mu}$ the mean over datapoints and $\vb{C}$ is the whitening transform. Let $\vb{X}$ be the $n \times d$ matrix consisting of $n$ $d$-dimensional centered inputs. 
ZCA uses
\begin{equation}
\vb{C} = \vb{\Lambda}^{-1/2} \vb{U}^\top
\end{equation}
with $\vb{U}\vb{\Lambda}\vb{U}^\top$ the eigendecomposition of the covariance matrix $\vb{\Sigma}=\vb{X}\vb{X}^\top/n$. 
\citet{Desjardins2015} propose to periodically (during training) measure the correlation, $\vb{C}$, at every layer of a deep neural network and to then apply eigendecomposition method to compute the ZCA transform.

This approach to achieving an exact whitening transform is an alternative to what we attempt in this work but has a number of drawbacks.
First, the complexity of computing the eigendecomposition itself has a cost equivalent to a matrix multiplication - meaning that this step alone is equally expensive to our decorrelation method.
Second, atop this decomposition, \citet{Desjardins2015} further compute a matrix inverse of this transform which allows them to periodically compute an exact ZCA-based decorrelation within each layer of their network while also undoing its impact on the layer's computation (by modifying the layer's weight matrix $\vb{W}$) to ensure that the newly introduced transform does not change the network's output.
We avoid the second step by simultaneously optimising for both the decorrelation and task at every layer and at every update step, without directly computing the ZCA transform.
This provides the benefit of not requiring any matrix inversion computation and also ensures that we are continuously optimising the decorrelation transform rather than doing so at fixed intervals.

\section{Impact of subsampling}
\label{sec:sampling}

\begin{figure*}[!ht]
\begin{center}
\includegraphics[width=0.9\linewidth]{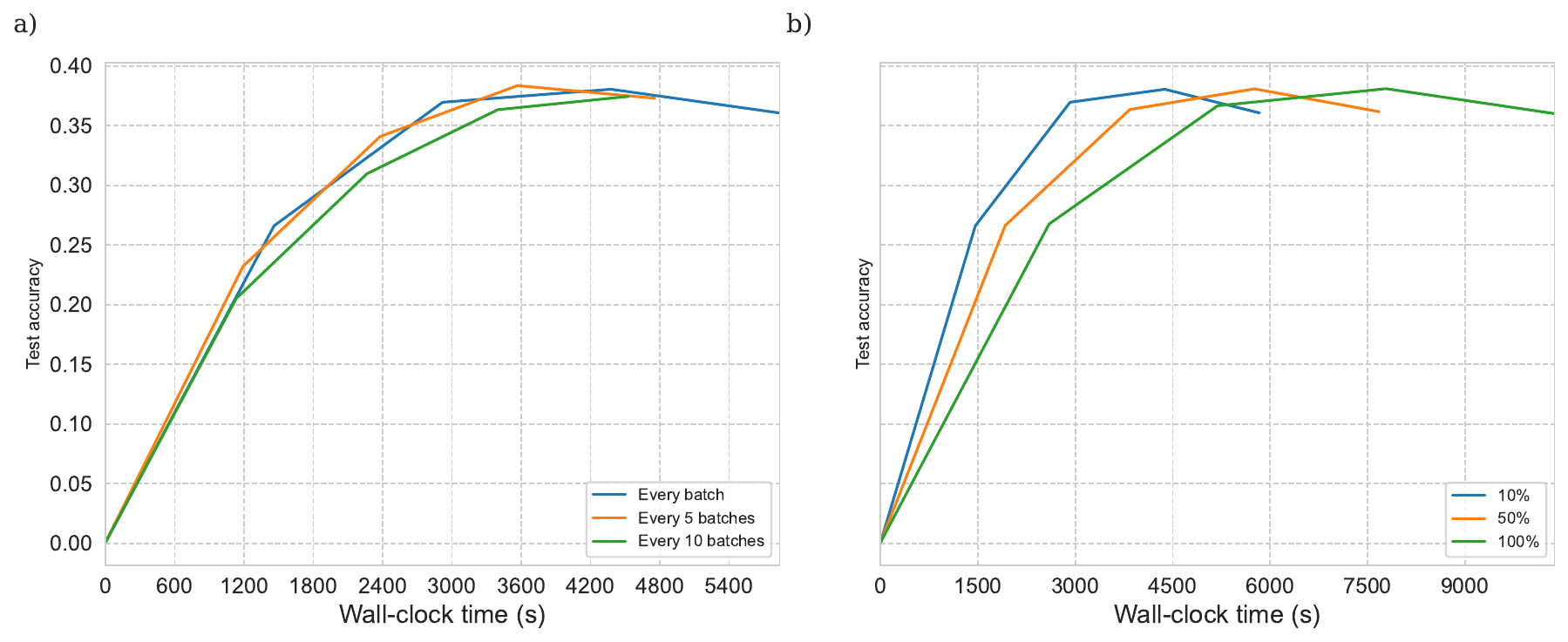}
 \caption{AlexNet test accuracy on ImageNet  a) Sampling every batch, 5 batches, and 10 batches. b) Sampling 100\%, 50\%, and 10\% of the batch.}
\label{fig:sampling}
\end{center}
\end{figure*}

Figure~\ref{fig:sampling} shows test performance as a function of wall-clock time for AlexNet trained on ImageNet at different sampling frequencies and sampling rates. Sampling 10\% of the batch appears sufficient for robust decorrelation while keeping computational overhead limited. Trying to save even more computation time by only updating the decorrelation matrix every 5 or 10 batches does not appear to speed up convergence in this experiment. Based on these results, a sampling rate of 10\% was used in all experiments, while the sampling frequency was not adjusted.

\section{Hyper-parameter optimization}
\label{sec:grid}

\begin{figure}[!ht]
\begin{center}
\includegraphics[width=0.8\textwidth]{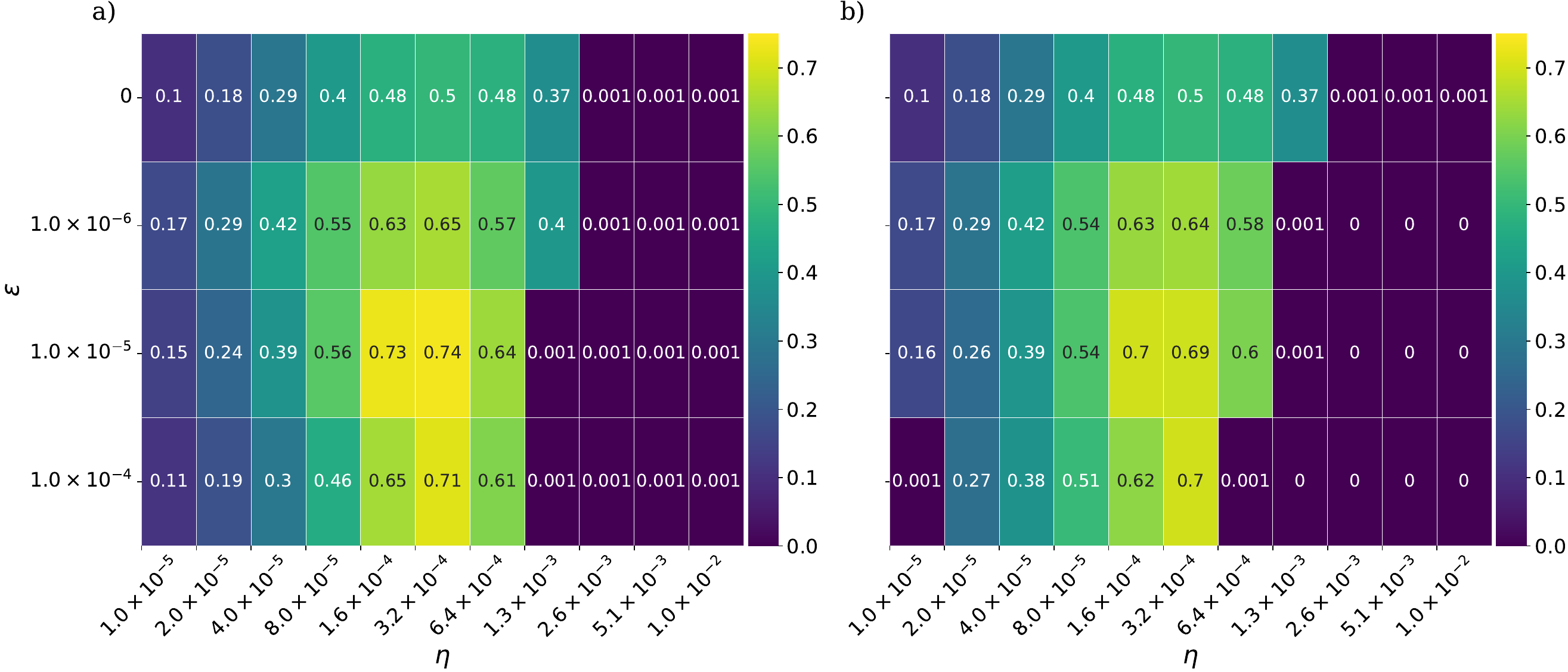}
 \caption{Train accuracy after five epochs for DBP when varying the BP learning rate $\eta$ and decorrelation learning rate $\epsilon$. Note that a decorrelation learning rate of zero corresponds to regular BP. a) Results using decorrelation ($\kappa=0$). b) Results using whitening ($\kappa=0.5$).} 
\label{fig:grid}
\end{center}
\end{figure}

A grid search across learning rate hyper-parameters was performed to ensure that each algorithm reached its optimal performance.
An exploratory analysis revealed that although $3.2 \cdot 10^{-4}$ gave slightly better performance after five epochs for BP and DBP, later in training a learning rate of $1.6 \cdot 10^{-4}$ was better. Therefore, the latter was chosen for all algorithms, which had the convenient side effect of using the same learning rate for all algorithms, making comparison more straightforward. Figure~\ref{fig:grid} depicts the grid search results for the ResNet18 model.

\section{Impact of whitening for ConvNets on CIFAR-10}
\label{sec:kappaeffect}

\begin{figure*}[!ht]
\begin{center}
\includegraphics[width=.8\linewidth]{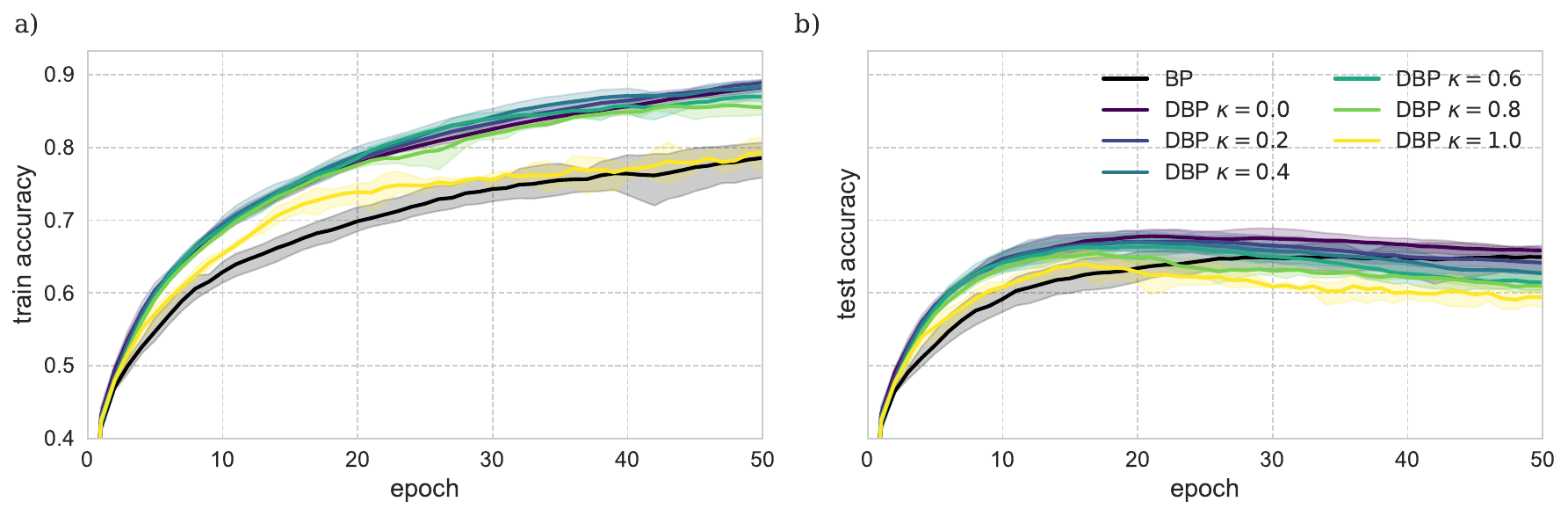}
 \caption{Performance on CIFAR10 for BP and several $\kappa$ settings of DBP. a) Train accuracy. b) Test accuracy.}
\label{fig:cifar}
\end{center}
\end{figure*}

To investigate the effect of the $\kappa$ parameter in a different setting than large (residual) DNNs, a small three-layer ConvNet was trained on the CIFAR-10~\citep{Krizhevsky} dataset. The model was trained using BP as welll as DBP under different values of the $\kappa$ parameter. As shown in Figure~\ref{fig:cifar}, a setting of $\kappa=0$ reached the highest peak test accuracy, though differences were small for for values less than $\kappa=0.6$. Note further that performance gains for this smaller model are even more pronounced than for the larger models. 

\end{document}